\documentclass[letterpaper, 10 pt, conference]{ieeeconf}  
\IEEEoverridecommandlockouts                              

\usepackage{microtype}
\usepackage{comment}

\usepackage{graphicx} 
\graphicspath{{./figures/}}
\usepackage{subcaption}
\usepackage{float}

\usepackage{amsmath} 
\usepackage{amssymb}  
\usepackage{units}
\usepackage{bm} 
\usepackage{multirow, array}

\usepackage{algorithm}
\usepackage{algpseudocode}

\makeatletter
\let\NAT@parse\undefined
\makeatother

\usepackage[square, numbers, sort&compress]{natbib}
\usepackage[hidelinks]{hyperref}
\usepackage[capitalise]{cleveref}

\crefname{equation}{}{}
\crefname{figure}{Fig.}{Figs.}

\usepackage{graphicx} 
\graphicspath{{./figures/}}
\usepackage{subcaption}
\usepackage{float}
\usepackage{xparse}
\usepackage{xcolor}
\usepackage{stfloats}
\usepackage{diagbox}
\usepackage{makecell}
\usepackage{tabularx}
\usepackage{tabulary}

{

\title{\LARGE \bf Sim-to-Real Learning for Humanoid Box Loco-Manipulation} 
\author{Jeremy Dao, Helei Duan, Alan Fern
\thanks{*This work is supported by the NSF Grant No. IIS-1849343, DGE-1314109, and DARPA Contract W911NF-16-1-0002.}
\thanks{All authors are with Collaborative Robotics and Intelligent Systems Institute, Oregon State University, Corvallis, Oregon, 97331, USA. }
\thanks{Email: \{\footnotesize daoje, duanh, afern\}@oregonstate.edu.}
}

\begin{document}

\maketitle
\thispagestyle{empty}
\pagestyle{empty}

\begin{abstract}
 In this work we propose a learning-based approach to box loco-manipulation for a humanoid robot. This is a particularly challenging problem due to the need for whole-body coordination in order to lift boxes of varying weight, position, and orientation while maintaining balance. To address this challenge, we present a sim-to-real reinforcement learning approach for training general box pickup and carrying skills for the bipedal robot Digit. Our reward functions are designed to produce the desired interactions with the box while also valuing balance and gait quality. We combine the learned skills into a full system for box loco-manipulation to achieve the task of moving boxes from one table to another with a variety of sizes, weights, and initial configurations. In addition to quantitative simulation results, we demonstrate successful sim-to-real transfer on the humanoid robot Digit. To our knowledge this is the first demonstration of a learned controller for such a task on real world hardware.



\end{abstract}

\section{Introduction}

Most reinforcement learning (RL) research for legged robots focuses on locomotion skills such as walking, running, and navigating various terrains \cite{Castillo2021, Tsounis2019, Li2021, Lee2020, Siekmann2020}. On the other hand, most RL research for object manipulation has focused on non-legged robots with fixed or stable bases (e.g. grounded robotic arms) \cite{Zhu2019, Wang2020, sun22a}. RL has been much less explored for loco-manipulation with legged robots, which involves both movement with and manipulation of objects. In particular, we are unaware of any work that has demonstrated successful sim-to-real RL for loco-manipulation with a bipedal humanoid robot. This is a particularly challenging problem due to the need for full-body coordination to manipulate and move objects while also maintaining balance.    


The main goal of this paper is to design and evaluate a sim-to-real RL approach for loco-manipulation with a real humanoid robot. In particular, our target task is to walk up to a box, stop, pick up the box, carry it to another table, and then set it down. Using data-driven RL for this purpose has the potential to facilitate generalizion to boxes of different sizes, masses, poses, and locations.  
Designing a sim-to-real RL solution, however, raises a question.
There is ambiguity about what is the ``best" or most desirable way to pickup and move a box. How should the hands move, and how should the overall motion of the body coordinate with the hands to produce smooth, non-agressive motion that is still robust? Furthermore, what kind of dense reward signal can we provide to faciliate learning such a motion? 

\begin{figure}
    \centering
    \includegraphics[width=0.99\columnwidth, trim={0cm 0cm 0cm 1.8cm}, clip]{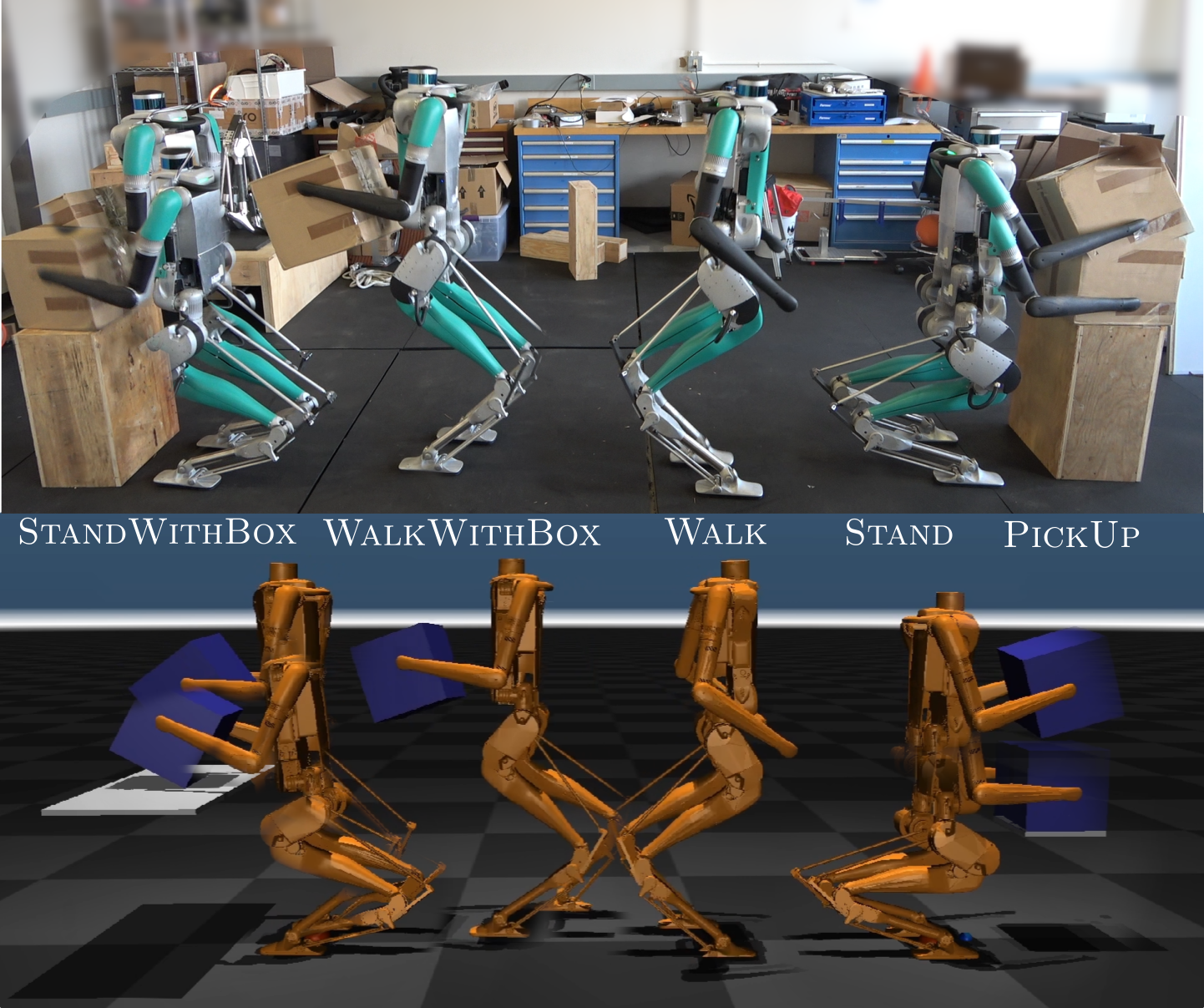}
    \caption{We learn box loco-manipulation policies in simulation and transfer directly to real hardware. We break the task down into 5 separate policies: \textsc{Walk}, \textsc{Stand}, \textsc{PickUp}, \textsc{WalkWithBox}, and \textsc{StandWithBox}.}
    \label{fig:hw_box}
\end{figure} 

This paper addresses this challenge by making the following contribtions:
\begin{itemize}
    \item We construct a reward function for learning a ``box pickup" skill and the associated locomotion skills. By designing these functions with box interaction in mind, specifying when the hands should move to contact the box, and when to apply forces to move the box, along with regulating the motion of the hands, we can learn smooth, gentle motions that still achieve the task.
    \item Compared to previous works on locomotion, we find that for box pickup using a ``relative" action space, where the policy outputs are PD setpoint commands to be added to the current joint position, is more suited for learning box pickup. This change in action space results in faster learning and less forceful interaction with the box.
    \item We present a full box loco-manipulation system where each skill, standing and walking with and without a box, and box pickup, is individually learned. We show successful sim-to-real on the humanoid robot Digit, which to our knowledge is the first demonstration of a learning controller achieving a loco-manipulation task on real world hardware.
\end{itemize}


\section{Related Works}

\subsection{Loco-Manipulation for Humanoids}
Much of prior research on loco-manipulation for humanoids has been largely model-based and focuses on mostly static manipulation tasks.
Model-based approaches have been explored \cite{Arisumi2007, Harada2005} for lifting up boxes, planning a whole-body motion and compensating for the extra force applied by the box. 
However, these methods produce very conservative motion to ensure absolute stability at all times, resulting in pickup motions that take over 20 seconds to be completed.
Planning based approaches \cite{Settimi2016, Ferrari2017, Karumanchi2017} have been explored for producing long loco-manipulation trajectories, with \cite{Settimi2016} even showing hardware results. 
Work focused on locomotion \cite{Harada2007, Sato2021} has explored how to modify a humanoid's gait when interacting with objects.
These strategies often augment some baseline gait controller to account for and modify the motion of the manipulated object.

Prior work that has used learning for humanoid or bi-manual manipulation tasks have been mostly trajectory based or rely on demonstrations as targets for imitation learning. For example, \citeauthor{Seo2023} (\citeyear{Seo2023}) uses teleoperation to obtain expert trajectories to imitate, while \citeauthor{Liu2023} (\citeyear{Liu2023}) uses motion tracked human demonstrations for a similar manipulation tracking task. This requirement of prior expert data can introduce another failure point in the learning system, as motion tracking error or alignment error between the human and robot data can cause componding issues down the line. 

\subsection{RL for Loco-Manipulation}
The use of learning for loco manipulation has been largely centered around mobile base or quadraped robots with an attached manipulator arm. 
Due to the inherent stability of such a system a common strategy is to separate out the locomotion and manipulation control in to two separate modules. \citeauthor{Ma2022} (\citeyear{Ma2022}) utilizes RL for only locomotion while the manipulator arm is controlled by a MPC planner, while \citeauthor{Sun2022} (\citeyear{Sun2022}) learns separate grasping and navigation policies. We utilize a similar separation of locomotion and manipulation skills, in that locomotion and box pickup are separate control policies, but unlike \citeauthor{Ma2022} (\citeyear{Ma2022}) our learned policies control all the joints, so that coordinated full body motion is always possible. 
\citeauthor{Fu2022} (\citeyear{Fu2022}) and \citeauthor{Arcari2022} (\citeyear{Arcari2022}) both learn a more unified control approach that commands both the legs and manipulator arm from the same controller. \citeauthor{Fu2022} (\citeyear{Fu2022}) learns a policy that executes base velocity and end effector position commands from a higher level planner, while \citeauthor{Arcari2022} (\citeyear{Arcari2022}) learns from a MPC trajectories.

Perhaps the most relevant prior work to our research is \citeauthor{Xie2023} (\citeyear{Xie2023}), which targets the same humanoid box pickup and locomotion task that we do. We follow a similar strategy of breaking down the task into different skills with a control policy for each. The main differences that we try to improve upon is to remove the reliance on trajectory tracking and focus more on hardware execution instead of an animation setting.


\section{System Overview} \label{sec:sys_overview}

\begin{figure}[t]
    \centering
    \includegraphics[width=0.95\columnwidth]{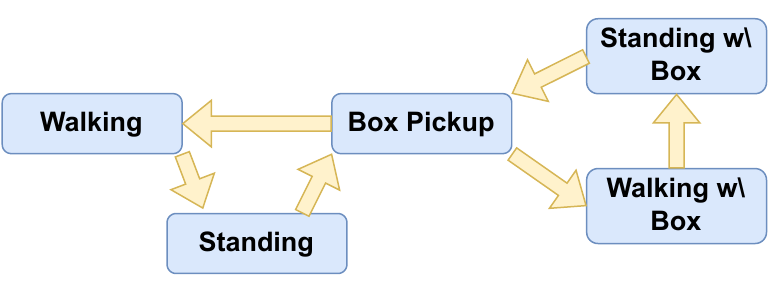}
    \caption{Allowed transitions between the 5 different policies. }
    \label{fig:skill_transition}
\end{figure}

We break down the full box loco-manipulation task into 5 distinct behaviors: 1) \textsc{Walk}, for basic walking, 2) \textsc{Stand}, for transitioning to a standing position after walking, 3) \textsc{PickUp}, for picking up the box, 4) \textsc{WalkWithBox}, for walking while holding a box, and 5) \textsc{StandWithBox} for transitioning to standing while holding a box. For each behavior we learn a distinct control policy, which has behavior specific command paramters. Box loco-manipulation involves only a subset of the possible transitions between these policies. For example, we will never go directly from walking to box pickup, and instead will always use standing as an inbetween behavior. \cref{fig:skill_transition} shows the allowed behavior transitions that we target for the box loco-manipulation task.



All of our policies follow the same general learning framework, with changes mainly to the reward function to learn different skills. 
Following prior work on bipedal locomotion \cite{Dao2022} we use an LSTM neural network with two 128-dimensional recurrent hidden layers to represent our policies, with the inputs and outputs adapted for the Digit humanoid robot, which has 20 DoFs and 10 unactuated joints.
The input to each policy is a pair $(s,c)$, where $s$ is the 67-dimensional robot-state vector and is the same for all policies, and $c$ contains a policy specific command. 
The policy runs at 50Hz and outputs actions corresponding to PD setpoint targets for each actuated joint. These are passed to a PD controller running at 2kHz with fixed gains.

We train policies in simulation using the MuJoCo physics engine \cite{todorov2012mujoco} and use the actor-critic PPO algorithm \cite{Schulman2017} with a clipped objective and gradient clipping. To facilitate sim-to-real transfer, we use dynamics randomization \cite{Peng2018} during training on parameters such as body mass, joint damping, body center of mass location, and friction. Ranges for these randomizations can be found in table \cref{table:DR}. Rather than use a state estimator during training and testing, we simply use the true simulation state and apply random noise on top of it. We train using 80 cores on a dual Intel Xeon Platinum 8280 server.


Each of our reward functions is defined in terms of a distinct set $\{(r_i,w_i)\}$ of paired reward terms and weights. The corresponding reward function is then defined by: 
\begin{align*}
R = \sum_i w_i\text{exp}(-r_i)
\end{align*}
In most cases, each of the reward terms will represent some cost to be avoided, so that the negative exponential will be smaller for larger costs. The state variables used below to define the reward functions are defined in \cref{table:rew_def}.

\begin{table}[t]
\vspace{0.25cm}
\centering
 \begin{tabular}{|c | c|} 
 \hline
 Body Mass & $[-0.2, 0.2]$\\ 
 \hline
 Joint Damping & $[-0.5, 2.5]$\\ 
 \hline
 Ground Friction & $[-0.3, 0.2]$\\ 
 \hline
 Center of Mass Position & $[-0.05, 0.05]$\\ 
 \hline
\end{tabular}
\caption{Dynamics randomization ranges. All ranges are in percentage values, so a range of $[-0.2, 0.2]$ corresponds to a range of $\pm20\%$ of the original value.}
\label{table:DR}
\end{table}

\section{Learning Loco-Manipulation Skills}
 

\begin{table}[t]
\vspace{0.25cm}
\centering
 \begin{tabular}{|c | c|| c | c|} 
 \hline
 $p_\text{x,t}$ & position of x at time t&  $p_\text{x,t}^*$ & \makecell{target position of x\\ at time t}\\ 
 \hline
 $F_\text{x, y}$ & \makecell{contact force \\between x and y}& $c_\text{x, y}$ & \makecell{contact indicator\\ between x and y \\(1 if in contact,\\0 otherwise)}\\ 
 \hline
 $\Phi_\text{x}$ & roll angle of x &  $\Theta_\text{x}$ & pitch angle of x\\ 
 \hline
 $v_\text{x}$ & velocity of x &  $a_\text{x}$ & acceleration of x\\ 
 \hline
 $q_\text{x}^*$ & \makecell{quaternion \\orientation of x} &  $p^{i}_{x/y/z}$ & x/y/z position of i\\ 
 \hline
 $u_t$ & action at time $t$ & $\tau$ & torque\\ 
 \hline
 
\end{tabular}
\caption{Definition of state variables used in reward functions.}
\label{table:rew_def}
\end{table}

\subsection{Box Pickup Policy} \label{sec:box_pickup}


We want to train a control policy that is capable of picking up boxes.  
It should be able to handle boxes of varying sizes and masses, along with the box starting in a range of positions in front the robot. In addition, we aim to learn pickup behavior that has smooth, natural motions and timings (e.g. not too fast or too slow). To define the pickup behavior we break it down into two phases: 1) a contact phase, where the hands aim to make contact with the box, and 2) the lift phase, where the hands aim to move the box to a target position. The phases are dictated by hand selected times, relative to the start of the behavior, which is time $t=0$. In particular, the contact phase ends at time $t_{\text{contact}}=100$ and then lift phase ends at $t_{\text{lift}}=175$. 

The command input $c$ to the pickup policy consists of: 1) box dimensions, mass, and starting pose, 2) the target position to move the box, and 3) two phase indicators $p_{\text{contact}}$ and $p_{\text{lift}}$, which each go linearly from 0 to 1 from the start to end of each phase.   
For ease of hardware testing, we assume that the box pose is only explicitly known at the start of the behavior. Once the box is picked up and has moved, we approximate the pose with the average pose of the hands, which avoids the need for real-time estimation of box pose. Given that the hands are in contact with the box, it is reasonable to assume that the box centroid position is going to be roughly between the hands. 

The action space of box pickup policy differs from the locomotion policies. In particular, the locomotion policies add the policy's action output to a static ``neutral" offset. Rather, for box pickup we found that adding the action output to the current motor positions resulted in faster learning (see \cref{sec:sim_results}). This is likely due to the much larger variance in motion profile of the pickup behavior from the neutral position compared to locomotion behavior.  


\noindent\textbf{Scenario Distribution.} 
At the beginning of each training episode, we spawn a box in a random location in front of the robot. The task is then to move the hands to the box, pick it up, and move it to a randomized target location above the table. 
The starting box location may be randomized within 35 to 50cm in front of the robot, $\pm30$cm to the side of the robot, and be on the ground or up to 1.3m in the air. The starting box yaw orientation may be $\pm22.5^\circ$. The target location is randomized within the same range, except the z position of the target will only be above the starting location. 
The box length, width, and height may each be between 20 to 45cm, and the mass may be between 1 to 10kg. 

\noindent\textbf{Reward Function.}
To train box pickup policies, we take inspiration from previous work on learning locomotion policies \cite{Siekmann2020, Dao2022}. 
Those works described walking through stance and swing phases, in other words specifying how and when the feet should contact with the ground. 
This general principal of perscribing contact can be applied to almost any interaction task, and is the main idea we utilize for our reward function. In contrast to locomotion, box pickup is a non-periodic behavior and we use a non-periodic clock signal with phase reference points $t_{\text{contact}}$ and $t_{\text{lift}}$ to define the reward. 


In particular, the reward function has the form
\[
R = R_{\text{traj}} + R_{\text{box}} + R_{\text{stand}} + R_{\text{reg}}
\]
where $R_{\text{traj}}$ rewards hand motion consistent with a box pickup, $R_{\text{box}}$ is designed to encourage proper interaction with the box, $R_{\text{stand}}$ aims to ensure the robot is stably standing, and $R_{\text{reg}}$ aims to regulate forces to ensure smoothness and help sim-to-real transfer. Below we describe the set of reward terms used to define each of these components noting that the weights for each term are provided in Table \ref{table:rew_weights}. 

\begin{figure}[t]
    \centering
    \includegraphics[width=0.95\columnwidth, trim={8.25cm 5cm 2cm 6.5cm}, clip]{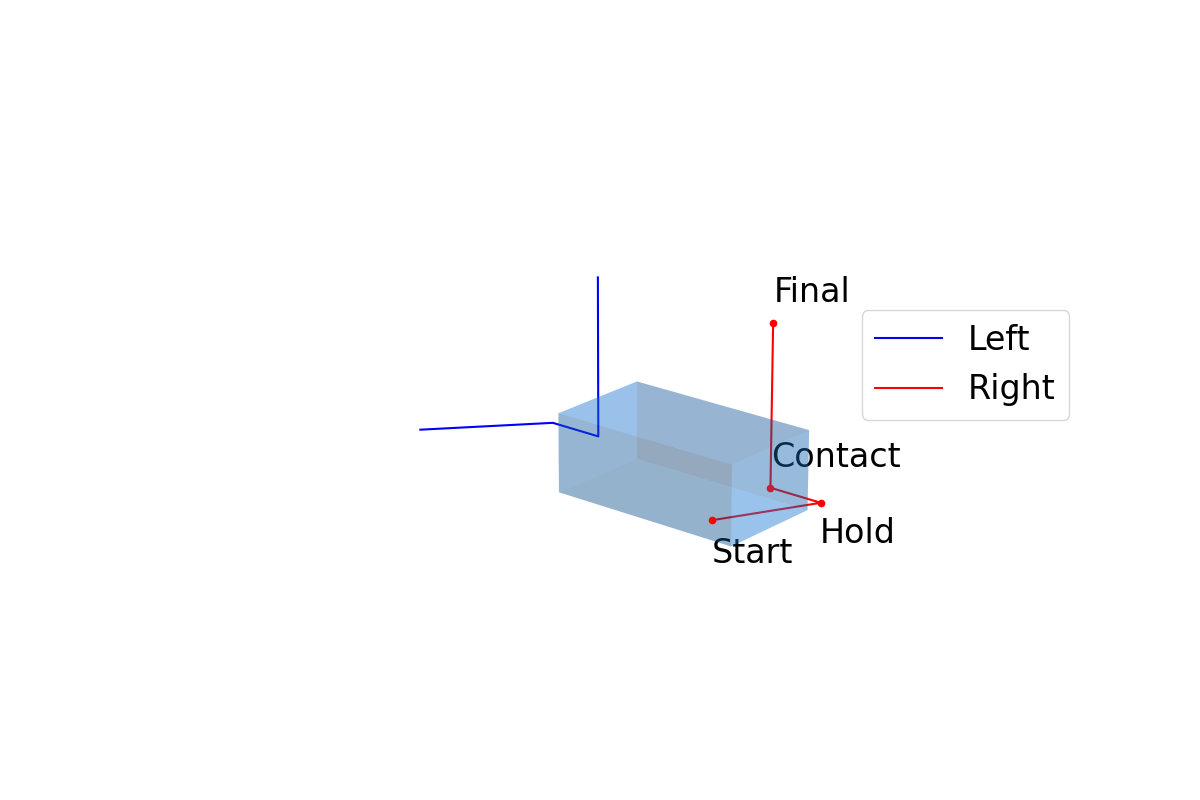}
    \caption{Example hand position trajectory for a box pickup. The hands start from the initial robot pose, move to the side of the box (shown in blue), make contact with it, and then bring it to the target location. In this example the target location is directly above the box.}
    \label{fig:handtraj}
\end{figure} 

$R_{\text{traj}}$ rewards being close to the target trajectory over time and having a hand orientation of zero using the following terms:
\begin{small}
\end{small}
\begin{align*}
    &r_{\text{hand\_pos}} = \|p_{\text{hand,t}}^{\text{left}} - p_{\text{hand,t}}^{*\text{left}}\| + \|p_{\text{hand,t}}^{\text{right}} - p_{\text{hand,t}}^{*\text{right}}\|\\
    &r_{\text{hand\_roll}} = |\Phi_{\text{hand}}^{\text{left}}| + |\Phi_{\text{hand}}^{\text{right}}|
\end{align*}
To describe the desired hand motions, we construct a hand position trajectory by choosing three goal hand positions, along with a timestamp for each, and linearly interpolate between them as shown in Figure \ref{fig:handtraj}. For the contact phase, the hands should first be out along each side of the box, and then eventually actually touching the box along it's transverse faces. 
Since $t_\text{contact}$ is 100 timesteps (2 seconds), we design the hands to be 10cm besides the box (goal position 1) at 1.5 seconds and then at 2 seconds the hands should be on the side faces of the box (goal position 2). The last goal postion 3 occurs at $t_\text{lift}$ (3.5 seconds) with the hands on either side of the target location, spaced evenly with a box's width distance between them.
We were also able to train policies without this hand position trajectory. However, we found that these policies trained significantly slower (see section VII) and had less success on hardware. 

\begin{small}
$R_\text{box}$ rewards contact with the box at the correct time (not too early), lifting it off the table, and moving the box to the target location with a flat orientation using the following terms: 
\end{small}
\begin{align*}
&r_\text{contact} = \ln{\left(0.05 \cdot c_{\text{hand,box}}^{\text{left}} + 0.05 \cdot c_{\text{hand,box}}^{\text{right}}\right)}\\
&r_{\text{box\_pos}} = \|p_{\text{box}} - p_{\text{box}}^*\|\\
&r_{\text{box\_orient}} = |\Phi_{\text{box}}| + |\Theta_{\text{box}}|\\
&r_{\text{table}} = \textbf{F}_{\text{table,box}}\\
\end{align*}
It is important to highlight the weighting for $r_\text{contact}$, weight $w_\text{contact} = I[t \geq t_{\text{contact}}]\cdot]$, where $I[\cdot]$ is the indicator function. This causes the contact reward term to only be active at the end of the first phase, which helps prevent early contact with the box. 

$R_\text{stand}$ rewards the robot for standing stably while picking up the box. 
This is done by keeping the center of pressure (CoP) in the middle of the support polygon (or the average foot location), the torso orientation upright and the feet orientation facing forward, motor velocity small (try to move as little as possible), and the feet velocities at zero:
\begin{align*}
    &r_{\text{CoP}} = \|p_\text{CoP} - p_\text{avg foot pos}\|\\
    &r_{\text{base\_orient}} = 1 - \left<q_{\text{torso}}, (1, 0, 0, 0)\right>^2\\
    &r_{\text{foot\_orient}} = (1 - \left<q_{\text{foot}}^{\text{left}}, (1, 0, 0, 0)\right>^2) + \\&\qquad\qquad\quad(1 - \left<q_{\text{foot}}^{\text{right}}, (1, 0, 0, 0)\right>^2)\\
    &r_{\text{motor\_vel}} = \|v_{\text{motor}}\|\\
    &r_{\text{feet\_vel}} = \|v_{\text{foot}}^{\text{left}}\| + \|v_{\text{foot}}^{\text{right}}\|
\end{align*}

Finally, $R_\text{reg}$ encourages smooth motions of both the robot and box itself. It also rewards gentle hand interaction with the box: 
\begin{align*}
    &r_{\text{action}} = \|u_{t} - u_{t-1}\|\\
    &r_{\text{torque}} = \|\tau\|\\
    &r_\text{hand\_force} = \|F_{\text{hand,box}}^{\text{left}}\| + \|F_{\text{hand,box}}^{\text{right}}\|\\
    &r_{\text{box\_acc}} = \|a_{\text{box}}\|
\end{align*}
The total reward $R$ also has two sparse penalties. There is a $-0.1$ penalty if there is a self-collision or if the velocity of either hand exceeds 1.0 m/s.
The weightings of each dense reward term used is in \cref{table:rew_weights}.
To encourage useful exploration, we also use the following termination conditions, which encourages learning critical aspects of the behavior before more detailed aspects.  
\begin{itemize}
    \item If the height of the robot torso is less than 40cm.
    \item If the pitch angle of the robot torso is greater than $35^\circ$.
    \item If either foot loses contact with the ground.
    \item If the robot makes contact with the table holding up the box.
    \item If the box is on the ground.
    \item If it has been 0.5 seconds after the contact countdown and the hands are not in contact with the box.
    \item If it has been 0.5 seconds after the pickup countdown and the box is still in contact with the table.
\end{itemize}

\begin{table}[h]
\centering
\resizebox{1\columnwidth}{!}{%
\begin{tabular}{l|l||l|l}
\hline
Name & Weight & Name & Weight \\ \hline
 Hand Position & 0.15 & Hand Roll & 0.05 \\ \hline
 Box Position & 0.15 & Box Orientation & 0.05 \\ \hline
 Table force & 0.05 & CoP & 0.1 \\ \hline
 Base Orientation & 0.05 & Foot Orientation & 0.1 \\ \hline
 Motor Velocity & 0.05 & Foot Velocity & 0.05 \\ \hline
 Hand Force & 0.05 & Box Acceleration & 0.05 \\ \hline
 Action Change & 0.05 & Torque & 0.05 \\ \hline
\end{tabular}}
\caption{Reward weightings for the box pickup reward.}
\label{table:rew_weights}
\end{table}

\noindent\textbf{Domain Randomization.}
In addition to the standard dynamics randomization we use for all policies, we also add randomization to the policy's estimate of the box mass ($\pm0.5$kg), dimensions ($\pm5$cm), and starting location ($\pm5$cm). 

\subsection{Locomotion Policies}
To learn locomotion skills we use the same learning setup as described in \cite{Siekmann2020, Dao2022} and we refer readers to those works for more details on the reward function. The only changes made for this work was an additional reward term on the hand positions, to keep them in front of the torso and equidistant from each other. 

The box locomotion policies follow largely the same learning setup as the regular locomotion policies, with a few tweaks to accomodate the box. Similar to the box pickup policy, the command input $c$ for the box locomotion policy contains the box dimensions, mass, and a height command as input. 
This height command specifies at what height the box should be held at while walking.

\noindent\textbf{Scenario Distribution.}
At the start of each training episode the robot is initialized to a random starting pose (see \cref{sec:skill trans} for more details), at a random phase in the walking cycle, and with a random command. The commanded x velocity may be randomized between $[-0.5, 1.0]$m/s, y velocity between $[-0.3, 0.3]$m/s, turning rate between $[-\pi/8, \pi/8]$rad/s, and box height between $[1.0, 1.3]$m. At a random time during the trajectory the commands with be randomized again. 

\noindent\textbf{Reward Function.} 
The reward function for learning box locomotion policies is the same as the regular locomotion policies \cite{Siekmann2020, Dao2022} with the addition of a few terms similar to the box pickup rewards. These terms form the box reward we denote as $R_\text{box}$:
\begin{align*}
    &r_\text{box height} = \|p_\text{box} - [0.4, 0, h_\text{cmd}]\|\\
    &r_\text{box orient} = |\Phi_\text{box}| + |\Theta_\text{box}|\\
    &r_\text{box force} = \|F_{\text{hand,box}}^{\text{left}}\| + \|F_{\text{hand,box}}^{\text{right}}\|\\
    &r_\text{hand roll} = |\Phi_{\text{hand}}^{\text{left}}| + |\Phi_{\text{hand}}^{\text{right}}|
\end{align*}
There is also an additional termination condition. The box locomotion policy will always be initialized with a box already in the hands, and it is not allowed to lose contact with the box else the training episode will terminate.

\noindent\textbf{Domain Randomization.}
The regular locomotion policy uses only the base dynamics randomization described in \cref{sec:sys_overview}. For the box locomotion policy, similar to the box pickup policies, there is randomization to the policy's box mass ($\pm0.5$kg) and dimension ($\pm5$cm) input.

\subsection{Standing Policies}
We want to train standing policies to transition from walking to static standing states. 
Since the standing policy only has the goal of standing still, there is no $c$ input to the standing policy and it only receives the robot state $s$.

The standing box policy has the same goal and learning setup as the normal standing policy, with only minor modifications to the reward. Since in this case there is a box that we would like to control, the command input $c$ for the standing box policy includes the box dimensions, mass, and a box height command.

\noindent\textbf{Scenario Distribution.}
At the start of each training episode the robot is initialized to a random walking state (see \cref{sec:skill trans} for more details). For the box standing policy the box height command is also randomized. 

\noindent\textbf{Reward Function.} 
The reward function consists of the following terms (in addition to $R_\text{stand}$ and the usual regulator reward $R_\text{reg}$):
\begin{align*}
    &r_\text{base vel} = \|v_\text{base}\|\\
    &r_\text{height} = |h_\text{base} - 0.9|\\
    &r_\text{arm} = \|p_\text{arm}^{\text{left}} - [0.15, 0.3, -0.1]\| + \\&\qquad\quad\|p_\text{arm}^{\text{right}} - [0.15, -0.3, -0.1]\|\\
    &r_{\text{stance width}} = |(py_\text{foot}^{\text{left}} - py_\text{foot}^{\text{right}}) - 0.385|\\
    &r_{\text{stance x}} = |px_\text{foot}^{\text{left}} - px_\text{foot}^{\text{right}}|
\end{align*}
Note that here $p_\text{arm}^\text{left\textbackslash right}$ denotes the position of the arms in the torso's frame. 
There is also a sparse reward penalty of $-0.1$ is there is a self collision.
The training episode terminates if the robot has fallen over or if either foot is off the ground after 100 policy timesteps.

The box standing policy follows the same reward function with the addition of the box reward $R_\text{box}$ described above.
The termination conditions are the same as well, with an additional termination if either hand loses contact with the box. 


\noindent\textbf{Domain Randomization.}
Like the locomotion policies, the regular standing policy use only the base dynamics randomization while the box standing policy has randomization to the box mass ($\pm0.5$kg) and dimension ($\pm5$cm) input.

\subsection{Skill Transitions} \label{sec:skill trans}

Once we have all 5 necessary box loco-manipulation skills, we need to connect them together in order to achieve the full task. Through out execution, we will transtion from one skill to another, and the policies need to be able to handle such a transition. 
\cref{fig:skill_transition} shows the allowed transitions between different policies. Using this we can create a workflow in which one policy is used to generate an initial state distribution with which we can use to train the next skill.

More specifically, we first train a regular walking policy. We then use this to generate many low-speed states from which we want to transition to standing from. These form the initial state distribution for training the standing policy. 
We can then use the standing policy to generate intial states for the box pickup policy and so on.


To ensure smooth transitions between policies, we also linearly interpolate between the two policies' actions over 10 timesteps. 
This action ``warmup" is applied during training as well, and so during initial state distribution generation we save the last applied action along with the robot state.

\section{Simulation Results} \label{sec:sim_results}

We first conduct quantitative evaluations for the Digit model in the MuJoCo simulator. Each evaluation averages over 10K episodes using the same episode generation approach as described for training. For example, for the evaluation of the \textsc{PickUp} policy we randomly sample box pose, mass, size, and target location. 

\textbf{Overall Policy Performance.} For each of the 5 learned policies we consider an episode a success if: 1) the robot does not fall over, and 2) the box remains in the robot's hands at the end of the episode for behaviors that involve the box. Further, for behaviors that involve the box, we also measure the error for successful episodes, which is the distance between the final position of the box and the commanded target position of the box.  
\cref{table:sim_table} records the success rate and error for each of the policies. Overall we see that the policies all achieve a high success rate and also produce relatively small errors. These results show that the policies are all quite robust to the randomization of the scenario distribution and the transition starting states that they will typically be initiated in. 

\textbf{Limits of PickUp Policy.} We also conducted additional simulation experiments for our most complex full-body behavior, PickUp. First, we evaluated its performance outside of the training distribution of box parameters. To do this we performed an incrementally increasing search for the maximum box mass, size, and starting pose that the policy can successfully pickup. In each test, only one parameter type is varied at a time with all other box parameters kept at the mean value of the training distribution. In particualar, we turned domain randomization off and continually incremented the particular parameter under test until a pickup failure occured. 
\cref{table:max-pickup} shows that the learned PickUp policy generalizes quite well outside of the training distribution and is able to pickup boxes much larger and heavier than those it saw during training. 


\begin{table}
\vspace{0.3cm}
\centering
 \begin{tabular}{c|c|c|c|c|c} 
 \hline
 & PickUp & Stand & StandBox & Walk & WalkBox \\
 \hline
 Success \% & 96.15\% & 100\% & 95.25\% & 96.1\% & 100\% \\ 
 \hline
 Error (cm) & 7.93 & N/A & 1.64 & N/A & 2.39 \\ 
 \hline
\end{tabular}
\caption{Success rate and accuaracy for the 5 behavior policies as evaluated over 10K episodes in the MuJoCo simulator. Each policy is evaluated in a random start state resulting from a successful run of its typical preceeding behavior (e.g. PickUp starts in a Stand end state). }
\label{table:sim_table}
\end{table}

\begin{table}
\vspace{0.3cm}
\centering
 \begin{tabular}{c|c|c} 
 \hline
 & Maximum & Training Range \\
 \hline
 Mass (kg) & 22.9 & [1, 10] \\ 
 \hline
 Size (cm)  & 64  & [20, 45]\\ 
 \hline
 X displacement (cm) & 56 & [35, 50]\\ 
 \hline
 Y displacement (cm) & 41 & [-30, 30]\\ 
 \hline
 Z displacement (cm) & 130 & [0,130]\\ 
 \hline
Rotation & $45^\circ$ & $[-22.5, 22.5]^\circ$\\
\hline
\label{table:max-pickup}
\end{tabular}
\caption{Test of the box PickUp policy on box parameters outside of the training distribution. Each row shows the maximum successful value for the policy for that box parameter along with the range used for training. }
\end{table}

\begin{figure}[t]
    \centering
    \includegraphics[width=0.95\columnwidth, trim={0cm 0cm 2cm 2cm}, clip]{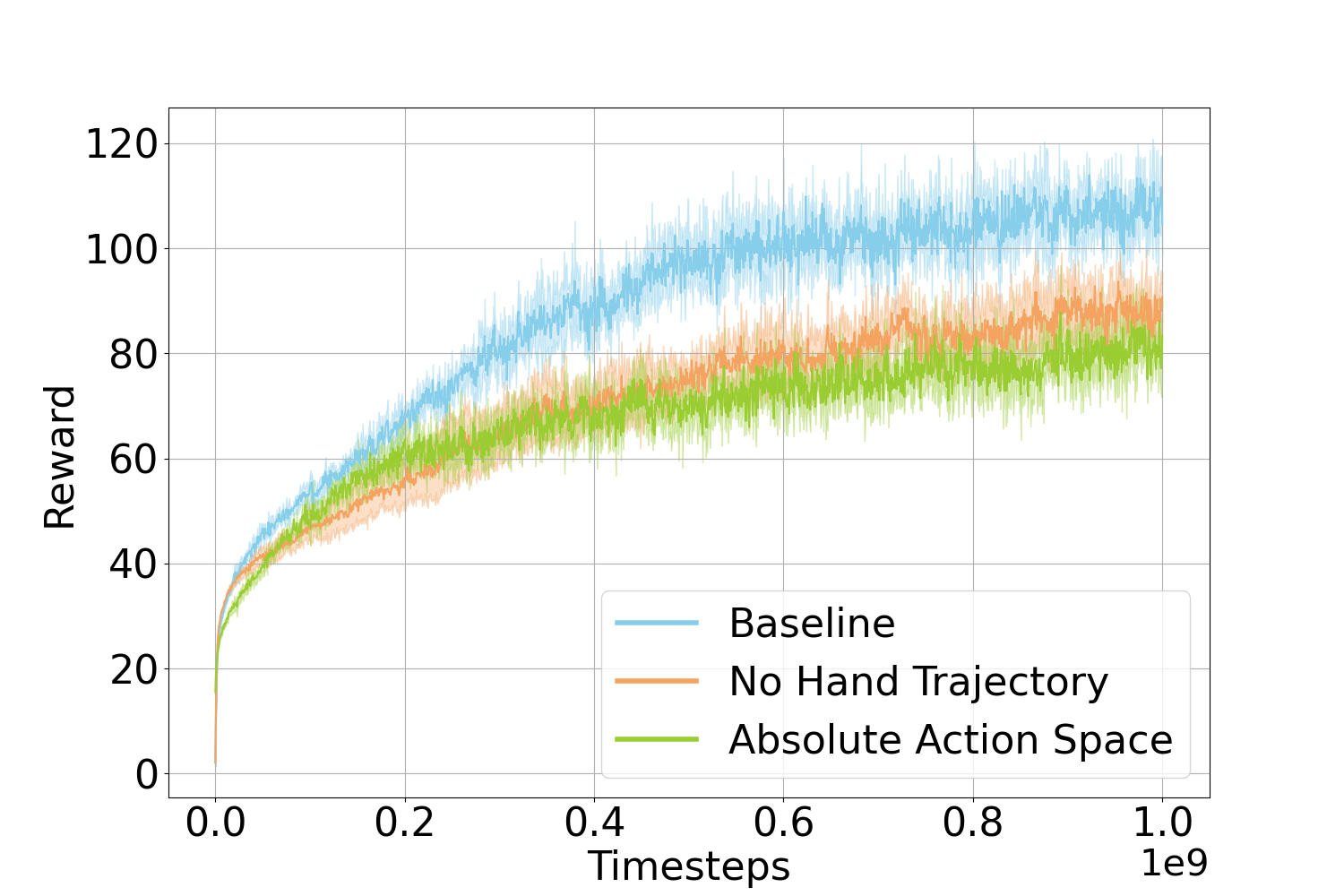}
    \caption{Reward curve comparison between different learning setups. ``Baseline" is the main learning setup we describe in \cref{sec:box_pickup}, ``No Hand Trajectory" is the same setup without the hand designed hand trajectory in the reward, and ``Absolute Action Space" adds the policy output to a fixed position offset rather than the current motor positions.}
    \label{fig:reward_comp}
\end{figure}

\textbf{PickUp Ablation.} We also perform some ablation studies on the learning setup with a focus on the PickUp behavior. To test if a simpler reward setup will still learn, we remove the hand trajectory tracking reward component. The target location for each hand $p^*$ is instead just the center of the left and right face of the box in the box target position, and does not change throughout the training episode. We found that we can still learn a successful box pickup policy, however, the sample efficiency suffers greatly. As can be seen in \cref{fig:reward_comp}, learning without the hand trajectory, approximately doubles the number of samples needed to reach the same reward as learning with the trajectory information.

Our PickUp policy uses a different action space than the locomotion and standing policies. Rather than adding the policy output to a fixed ``neutral" offset robot pose, the PickUp policy instead adds the policy output to the current joint motor positions. To test the effect of this change, we train policies with both action spaces. 
The reward curves for each are presented in \cref{fig:reward_comp}. As can be seen, using the fixed offset action space learns significantly slower, about as slow as not using the hand trajectory. We did not observe this same loss in sample efficiency for the standing and walking policies, leading us to believe that when the desired motion deviates greatly from the fixed position offset, it is beneficial to use a ``relative" action space instead.

\section{Hardware Results}

We show successful sim-to-real transfer for all trained policies, including transitions between each skill. We are able to perform box pickups for multiple boxes with different masses (from 1kg to 8kg) and sizes. The pickup will also work for boxes on the ground and those starting on a table. We refer readers to the attached video for hardware experiments.

To get the pose of the box during hardware experiments we utilize Digit's on board RGBD camera and an ArUco marker placed on the box. 
The box size and mass policy input is hardcoded beforehand. During hardware execution, skill transitions, navigation, and walking commands are handled by a human operator. 

Despite the success, we do observe sim-to-real issues on hardware. In particular, we observe policies leaning towards one direction during hardware execution. We suspect this is due to imu/oriention estimate differences on hardware. To counteract this, we offset the orientation input in the same direction as the observed leaning. 
While this ``center-of-mass trim" only needed turning once, it was policy specific.  We also found that the selected trim values did not work for the heaviest boxes (8kg), where we observed the policy leaning forward, requiring an increase in the trim. We hypthosize that this error might be caused by force or contact inconsistencies between simulation and hardware. 

\section{Conclusion}

We presented a learned system for performing box loco-manipulation on the Digit humanoid robot. We broke box loco-manipulation down into 5 different behaviors and learn separate policies for each. We were able to learn policies that achieve high performance in simulation for boxes of varying masses, sizes, and starting locations. We also demonstrated sucessful sim-to-real transfer of those polcies for full loco-manipulaiton episodes, involving walking to a box, pick it up, walk to a target location, and setting it down. To our knowledge this is the first such sim-to-real demonstration using fully learned policies. Future work will focus on making our system more autonomous by incorporating planning, navigation, and vision systems to remove the reliance on a human operator and ArUco markers. 
There is also room to improve our sim-to-real transfer to hardware. For this we will consider both enhancing the domain randomization using in simulation based training, and integrating real-world hardware data from successful and failed trials into the learning process.




\newpage
\addtolength{\textheight}{-7cm} 

\def\bibfont{\footnotesize}
\bibliographystyle{IEEEtranN}
\bibliography{references}

\end{document}